\documentclass[lettersize,journal]{IEEEtran}
\usepackage{amsmath,amsfonts}
\usepackage{array}
\usepackage[caption=false,font=normalsize,labelfont=sf,textfont=sf]{subfig}
\usepackage{textcomp}
\usepackage{stfloats}
\usepackage{url}
\usepackage{verbatim}
\usepackage{graphicx}
\usepackage{cite}
\usepackage{booktabs}
\usepackage{subcaption}
\usepackage{multirow}
\usepackage{threeparttable}
\usepackage{graphicx}
\usepackage{textcomp}
\usepackage{xcolor}
\usepackage{enumitem}
\usepackage{comment}
\usepackage{pifont}
\usepackage{amsthm}
\usepackage{algorithm}
\usepackage{hyperref} 
\usepackage{amsmath}
\usepackage{bm}
\usepackage{bbding}
\usepackage{amsthm}
\usepackage{algpseudocode}

\usepackage{fontawesome}

\newtheoremstyle{boldassumption}
  {\topsep}
  {\topsep}
  {\itshape}
  {}
  {\bfseries}
  {.}
  {.5em}
  {}

\theoremstyle{boldassumption}

\begin{document}

\title{UniFlow: A Foundation Model for Unified Urban Spatio-Temporal Flow  Prediction}

        
\author{Yuan~Yuan,
	Jingtao~Ding,
        Chonghua~Han,
        Zhi~Sheng,
        Depeng~Jin,~\IEEEmembership{Member,~IEEE}
        and~Yong~Li~\IEEEmembership{Senior Member,~IEEE}
\IEEEcompsocitemizethanks{\IEEEcompsocthanksitem Y. Yuan, J. Ding, C. Han, Z. Sheng, D. Jin, and Y. Li are with Beijing National Research Center for Information Science and Technology (BNRist), and Department of Electronic Engineering, Tsinghua University, Beijing 100084, China.
E-mail: y-yuan20@mails.tsinghua.edu.cn}
}



\maketitle

\begin{abstract}
Urban spatio-temporal flow prediction, encompassing traffic flows and crowd flows, is crucial for optimizing city infrastructure and managing traffic and emergency responses. 
Traditional approaches have relied on separate models tailored to either grid-based data, representing cities as uniform cells, or graph-based data, modeling cities as networks of nodes and edges. 
In this paper, we build UniFlow, a foundational model for general urban flow prediction that unifies both grid-based and graph-based data.  We first design a multi-view spatio-temporal patching mechanism to standardize different data into a consistent sequential format and then introduce a spatio-temporal transformer architecture to capture complex correlations and dynamics.
To leverage shared spatio-temporal patterns across different data types and facilitate effective cross-learning, we propose Spatio-Temporal Memory Retrieval Augmentation  (ST-MRA). 
By creating structured memory modules to store shared spatio-temporal patterns, ST-MRA enhances predictions through adaptive memory retrieval. 
Extensive experiments demonstrate that UniFlow outperforms existing models in both grid-based and graph-based flow prediction, excelling particularly in scenarios with limited data availability, showcasing its superior performance and broad applicability.
The datasets and code implementation have been released on \textcolor{blue}{\url{https://github.com/YuanYuan98/UniFlow}}.
\end{abstract}


\section{Introduction}

Urban flows refer to the dynamic movement patterns within a city, such as crowd flow and traffic flow~\cite{zheng2014urban}. 
Predicting these flows is essential for optimizing infrastructure and enabling efficient control and management of traffic, public transport, and emergency responses. 
To this end, various data collection approaches, including sensor networks~\cite{yu2018spatio}, GPS tracking~\cite{zhang2017deep}, and mobile networks~\cite{zhou2023towards, liyuan2023learning}, are widely employed.

Urban spatio-temporal flow prediction has been a long-standing research problem, attracting significant attention from researchers~\cite{wang2020deep, jin2023spatio, jin2023large, li2018diffusion,zhang2017deep,bai2020adaptive,geng2019spatiotemporal,yu2020spatio,chen2021s2tnet}. 
The data collected for this purpose are typically organized into two categories: grid-based and graph-based data. Grid-based data represents the city as a grid of uniform cells, each containing aggregated flow information, making it suitable for capturing local dynamics and interactions. 
Various models, primarily employing Convolutional Neural Network (CNN)-based techniques, have been proposed to handle grid-based data~\cite{zhang2017deep, liang2019urbanfm,liu2018attentive}. 
In contrast, graph-based data models the city as a network of nodes (e.g., intersections) and edges (e.g., roads), which better represents connectivity across the city. Numerous Graph Neural Network (GNN)-based models have been developed for prediction in this context~\cite{pan2019urban, yu2018spatio, wu2019graph, bai2020adaptive,liang2018geoman}.

Although existing works have made significant progress in this task due to the development of deep learning techniques,  they still focus on developing and training specialized models for different types of data. In this work, we aim to answer a crucial research question: \textit{\textbf{Is it possible to build a foundation model to unify urban spatio-temporal flow prediction? }} By doing so,  we can simplify this classical urban task by building a single, one-for-all model without the need for multiple specialized models.
The intuition behind this vision lies in the intrinsic connection between different data types, both representing the spatio-temporal rhythms of urban life and reflecting various facets of human activity. Grid-based data capture local, regular spatial relationships, while graph-based data describe complex, irregular connections. These aspects are interconnected; for example, traffic flow involves both regular spatial relationships (grid) and complex congestion connections between roads (graph). Moreover, universal spatio-temporal laws, such as periodicity, are not constrained by specific data structures. Addressing these data types separately hinders the ability to fully capture the spatio-temporal dynamics of a city.

\begin{figure}[t!]
    \centering
    \includegraphics[width=0.99\linewidth]{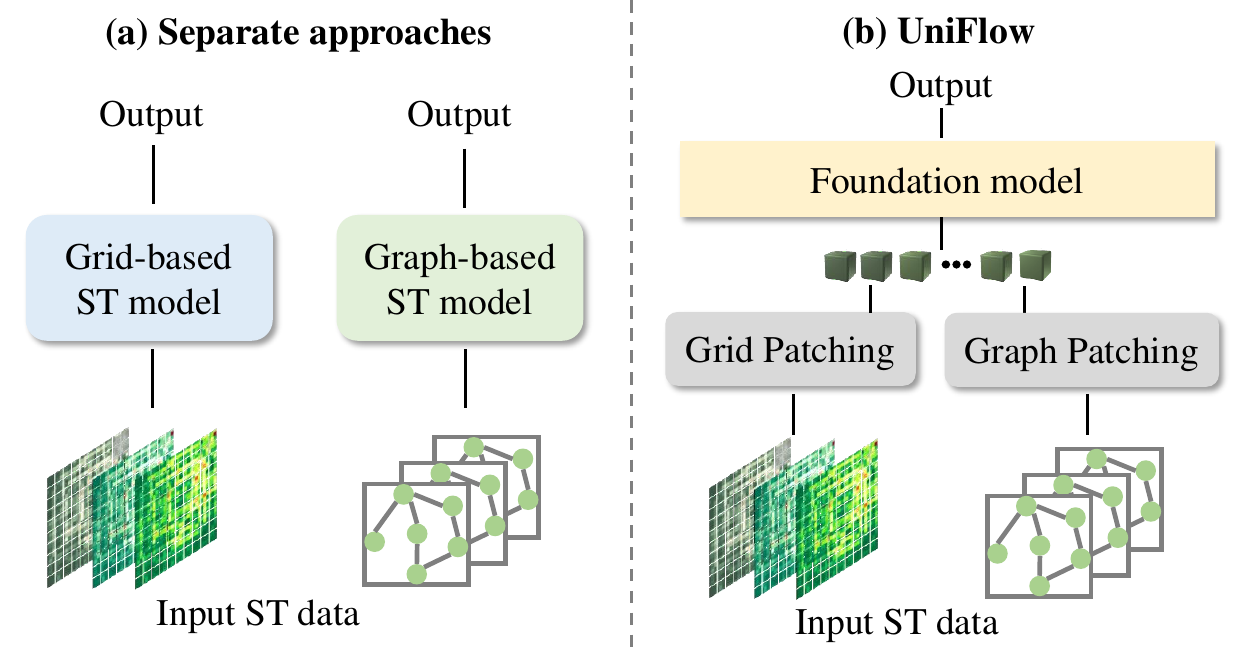}
    \caption{\textcolor{black}{High-level comparison between our proposed
model and separate modeling for spatio-temporal data. Our solution unifies grid and graph data formats.}}\label{fig:intro_fig}
\end{figure}

However, achieving such a unified foundation model is not a trivial task. 
The first challenge lies in integrating different data structures. Grid-based data, with its uniform distribution and local flow information, contrasts sharply with the network-like structure of graph-based data, which emphasizes connectivity. Unifying these different types of data into a cohesive model requires bridging their structural differences.
The second challenge is extracting and utilizing the underlying commonalities between these disparate data types, as they capture different aspects of spatio-temporal dynamics.
Directly applying a single model to fit all data without prior spatio-temporal knowledge makes it challenging to leverage shared commonalities, potentially undermining the objective of creating a truly unified model.

To address these challenges, we have developed UniFlow, a foundation model for unified urban  spatio-temporal flow prediction that handles both gird-based and graph-based data. 
First, we design a multi-view spatio-temporal patching mechanism to process different types of input data in parallel, unifying them into a consistent sequential structure.
Next, we develop a spatio-temporal transformer architecture to capture complex dynamics over time and space.
To extract fundamental spatio-temporal patterns independent of specific data and facilitate cross-learning across different data types, we propose Spatio-Temporal Memory Retrieval Augmentation (ST-MRA). The core idea is to generate prompts that are most beneficial for the different spatio-temporal patterns encountered. Leveraging well-established spatio-temporal domain knowledge, we create explicit and structured memory modules to store spatio-temporal patterns. The transformer is enhanced by dynamically interacting with this memory, improving its utilization of stored knowledge.
For each input sample, this process involves vectorizing queries based on extracted spatio-temporal patterns, retrieving relevant content from the memory, and generating personalized prompt vectors to guide the model's predictions. Unlike the non-parametric Retrieval-Augmented Generation (RAG)~\cite{lewis2020retrieval} approach widely used in large language models, ST-MRA is fully learnable, with both the memory modules and retrieval process being trainable.
In summary, our contributions are as follows:

\begin{itemize}[leftmargin=*]
    \item We pioneer the unification of different types of urban flow data into a single model for universal prediction.
    \item We propose UniFlow, a transformer-based unified model that leverages multi-view spatio-temporal patching and spatio-temporal memory retrieval augmentation to capture and utilize complex  spatio-temporal dynamics in urban flow data.
    \item Extensive experiments on nine real-world datasets demonstrate the superiority of UniFlow. Our one-for-all foundation model consistently  outperforms specialized, individually trained baseline models, achieving an average performance improvement exceeding 10\%. It highlights the significant potential of foundation models in this field.
\end{itemize}
\section{Related Work}

\subsection{Urban Spatio-Temporal Prediction}
Urban spatio-temporal flow prediction~\cite{wang2020deep,zheng2014urban} aims to model and forecast the dynamic patterns of urban activities over space and time, such as crowd flows and traffic flows. It provides crucial insights for urban decision-makers to optimize infrastructure and improve public services. Generally, these predictions rely on two types of data: (i) grid-based flow data and (ii) graph-based flow data. 
Deep learning methods have  achieved significant advancements in this task~\cite{wang2020deep, jin2023spatio, jin2023large, yuan2024spatiotemporal,yuan2023spatio}. 

For grid-based flow data, CNNs~\cite{li2018diffusion,zhang2017deep,liu2018attentive} and ResNets~\cite{zhang2017deep} have been introduced to model the regular adjacency spatial correlations. For graph-based flow data, various GNNs~\cite{zhao2019t, bai2020adaptive,geng2019spatiotemporal, jin2023spatio, jin2023transferable} are utilized to capture more complicated spatial relationships.  Besides, RNNs~\cite{wang2017predrnn, wang2018predrnn++,lin2020self} and transformers~\cite{chen2022bidirectional,jiang2023pdformer,yu2020spatio,chen2021s2tnet} are also utilized to model temporal dynamics. 
However,  most approaches remain constrained by the need to train separate models for different datasets.  In contrast, our proposed model allows to train across different data types  and provides a unified solution. 
\textcolor{black}{While transformer-based approaches~\cite{jiang2023pdformer} can, in theory, handle both grid and graph data, they face scalability issues when applied to large-scale graphs or grid with massive nodes or locations. }

\subsection{Foundation Models for Spatio-Temporal Data}
Building on the significant advancements in foundation models for NLP~\cite{touvron2023llama,brown2020language} and CV~\cite{rombach2022high,bai2023sequential}, recent efforts have introduced foundation models for urban tasks.
These research have explored the potential of large language models (LLMs) in this domain. Intelligent urban systems like CityGPT~\cite{feng2014citygpt,xu2023urban}, CityBench~\cite{feng2014citybench}, and UrbanGPT~\cite{li2024urbangpt} have shown their proficiency in handling language-based urban reasoning tasks.
There has also been great progress in foundation models for time series prediction~\cite{jin2023large,cao2023tempo,jin2023time, zhou2023one}. 
Spatio-temporal data, unlike the sequential nature of time series data, involves a more intricate setup due to its dependencies across space and time. The complexity deepens with different types of spatial correlations, including those arranged in structured grids and those mapped onto unstructured graphs, adding layers of difficulty to the modeling process.
UniST~\cite{yuan2024unist} provides a universal solution for multiple spatio-temporal scenarios but is still limited to grid-based data.
In Table~\ref{tbl:compare}, we contrast the key characteristics of  UniFlow against other related works.
UniFlow stands out as a universal model proficient in handling a diverse range of data types. Additionally, it excels in both few-shot and zero-shot learning scenarios.

\begin{table}[t!]
\caption{Comparison between existing works and UniFlow from
four aspects. }\label{tbl:compare}
\begin{threeparttable}
\resizebox{\columnwidth}{!}{
\begin{tabular}{cccc}
\hline
Approach & Model\tnote{(1)}  & Data\tnote{(2)}   & \textcolor{black}{Zero-shot} \\ \hline
TrafficBERT~\cite{jin2021trafficbert} & S & Graph-O   & $\times$  \\
GPT-ST~\cite{li2023generative} & S &  Graph-O  &$\times$ \\
ST-SSL~\cite{ji2023spatio} & S & Graph-O  &$\times$ \\
STEP~\cite{ShaoZWX22} & S &  Graph-O  & $\times$  \\
STG-LLM~\cite{liu2024large} & S & Graph-O    & $\times$  \\ 
UrbanGPT~\cite{li2024urbangpt} & U & Grid-M  &  \checkmark \\
PromptST~\cite{zhang2023promptst} & S & Grid-O  & $\times$ \\
UniST~\cite{liu2024large} & U & Grid-M  & \checkmark  \\\hline
UniFlow & U &  Graph-Grid-M  & \checkmark   \\ \hline  
\end{tabular}}
\begin{tablenotes}
    \footnotesize
    \item[(1)] S: specific model; U: universal model.
    \item[(2)] Data diversity: O denotes one dataset and M denotes multiple datasets.
\end{tablenotes}
\end{threeparttable}
\end{table}

\subsection{Retrieval-Augmented Generation} Retrieval-Augmented Generation (RAG)~\cite{lewis2020retrieval} enhances model performance by incorporating external knowledge into the generation process. Initially popularized in the context of natural language processing~\cite{gao2023retrieval, cheng2024lift}, RAG has shown promise in improving the quality and relevance of generated text by retrieving pertinent information from a vast corpus of documents. Recently, RAG techniques are also utilized in the recommender system~\cite{lin2024rella}, which designs semantic user behavior retrieval to improve the data quality. Applying RAG to spatio-temporal data presents unique opportunities but lacks explorations. In this work, inspired by the principles of RAG, we propose to utilize retrieval-augmented spatio-temporal knowledge.  In this way, we can enhance the model's ability to draw on relevant spatio-temporal patterns, thereby improving prediction accuracy and robustness.

\begin{figure*}[t!]
    \centering
    \includegraphics[width=\linewidth]{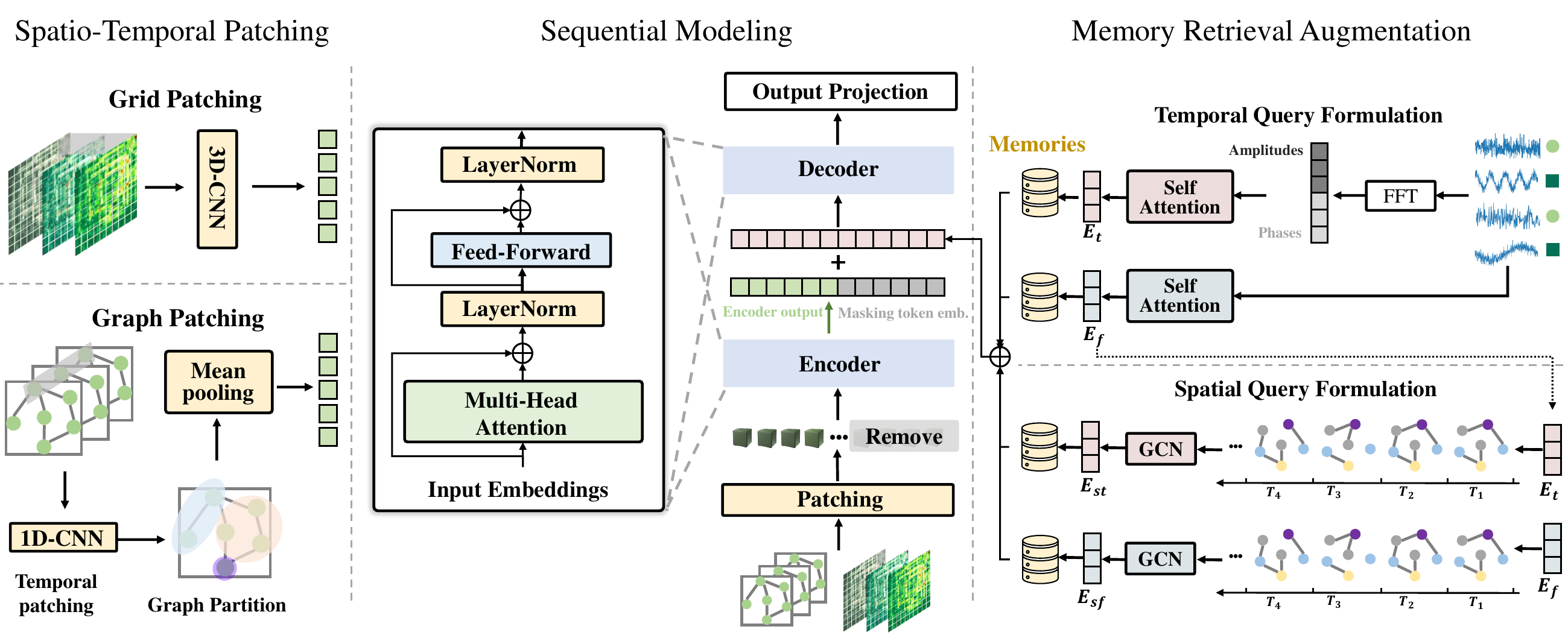}
    \caption{The overall architecture of UniFlow. For the input data, we first perform spatio-temporal patching to construct sequential structures. Then the transformer model performs sequential modeling, which is prompted by spatio-temporal memory retrieval augmentation.}
    \label{fig:model}
\end{figure*}

\section{Preliminary}

\subsubsection*{\textbf{Flow Data}} Spatio-temporal Flow data, denoted as \(X\), can be defined as a tensor with dimensions \(T \times N \times C\), where \(T\) represents the temporal period length, \(N\) is the number of locations, and \(C\) is the number of variables. Typically, urban spatio-temporal flow data are organized in two main types: grid-based and graph-based. 

For the grid-based data, \(X\) can be expressed as \(T \times H \times W \times C\), where \(H\) and \(W\) represent the longitude and latitude partitions, respectively, and \(H \times W = N\). 

For the graph-based data, \(X\) is represented as \(T \times N \times C\), where \(N\) corresponds to the number of nodes in the topology. Additionally, graph-based data includes a spatial topology denoted as $\mathcal{G} = (\mathcal{V},\mathcal{E}, \mathcal{A})$,  where \(\mathcal{V} = (v_1, v_2, ..., v_N)\) is the node set, \(\mathcal{E}=\{e_{ij}\}\) is the edge set, and \(\mathcal{A}=\{a_{ij} \in \{0, 1\}\}\in \mathbb{R}^{N\times N}\) denotes the adjacency matrix of the spatial graph. In this context, $a_{ij}=1$ indicates an edge between node $i$ and node $j$, while $a_{ij}=0$ indicates no edge.

\subsubsection*{\textbf{Spatio-Temporal Flow Prediction}} A common scenario involves using $H$ historical records to predict future $P$ flows.  Formally, given a set of historical spatio-temporal observations $\{X_t\}^{H}_{t=1}$, the objective is to predict the future spatio-temporal states $\{X_t\}^{H+P}_{t=H+1}$, where $H$ denotes the number of history time steps and $P$ is the  number of prediction time steps.  This involves learning a mapping $\mathcal{F}: \{X_t\}^{H}_{t=1} \rightarrow \{X_t\}^{H+P}_{t=H+1}$, which can be represented as:

\begin{equation}
    \hat{X}_{H+1:H+P} = \mathcal{F}(X_{1:H}),
\end{equation}

\noindent where \(\hat{X}_{H+1:H+P}\) denotes the predicted values  for the prediction period, and \(X_{1:H}\) represents the historical records. The goal is to predict future data points from  \(H+1\) to \(H+P\) based on the historical data spanning the period \(1\) to \(H\).

\section{Methodology}

Figure~\ref{fig:model} illustrates the overall architecture of UniFlow, which consists of three main components: (i) spatio-temporal patching, which converts spatio-temporal flow data with varied types into a standard sequential format, (ii) sequential modeling with transformers to process the patched spatio-temporal input, and (iii) spatio-temporal memory retrieval augmentation, which retrieves most useful prompts from shared memories to further enhance prediction performance.

\subsection{Spatio-Temporal Patching}

While the temporal dimension of flow data naturally exhibits a sequential structure between consecutive timestamps, the spatial organization does not inherently follow a sequential structure. Considering the intertwined nature of spatio-temporal dynamics, relying solely on temporal correlations limits the ability of transformers to fully capture spatio-temporal correlations. Therefore, implementing spatio-temporal transformers is necessary.

To transform flow data into a sequential structure, we design separate lightweight patching modules to patch flow data for both grid-based and graph-based types. Notably, for both types, we adopt a channel-independence strategy following existing practices~\cite{nie2022time}.
 This approach treats the flow data denoted as \(T\times N \times C\) as $C$  individual \(T\times N\) data sequences.  Thus, our goal is to transform the input with shape \(T \times N\) into a sequence with length of \(L\). 

 \subsubsection{\textbf{Spatio-Temporal Grid Patching}}
Grid-based flow data can be represented as  \(X\in \mathbb{R}^{T\times H \times W}\) with \(H\times W = N \text{ spatial regions}\). 
To convert this data into a sequential structure, we first use a three-dimensional convolutional neural network (3D-CNN) with kernel and stride sizes equal to the patch size $(p_t, p_s, p_s)$.
In this way, we divide the data into $L$ patches, i.e., $L = \frac{T}{p_t} \times \frac{H}{p_s} \times \frac{W}{p_s}$. We then arrange these smaller patches in a scan-line order, treating them as a sequence. The process can be formulated as follows:

\begin{align}
    & X' = \textsc{Conv3D}(X, \textsc{kernel}=(p_t,p_s,p_s), \textsc{stride}=(p_t,p_s,p_s)) ,\nonumber \\
    & S  = \textsc{Flatten}(X') . \nonumber
\end{align}

\noindent Thus, \(S\in \mathbb{R}^{L \times D}\) is a sequence of patches that can be fed into the transformer model for further processing. $D$ is the output dimension of the CNN model.

\subsubsection{\textbf{Spatio-Temporal Graph Patching}}
Graph-based flow data is represented as \(X\in \mathbb{R}^{T\times N}\), together with a graph topology \(\mathcal{G} = (\mathcal{V}, \mathcal{E})\), where \(\mathcal{V}\) is the set of nodes and \(\mathcal{E}\) is the set of edges. 
Unlike the grid-based data, graph topology lacks a regular structure, making CNN-based patching approaches inapplicable. 
To tackle this issue, we use separate patching modules to patch graph-based data. 
For the temporal dimension, we still use a 1D-CNN to extract features from the time series data.
To patch the spatial graph, we divide the overall graph into multiple subgraphs, each containing a set of nodes and edges. We adopt the METIS algorithm~\cite{karypis1997metis} to partition the graph into $L$ subgraphs.
The primary goal of METIS is to divide a graph into smaller parts while minimizing the number of edges that are cut and balancing the sizes of the parts. 

After dividing the graph into multiple subgraphs, we treat each subgraph as a patch. We obtain the patching results by averaging the pooled temporal features of different nodes within each subgraph. The graph patching process is formulated as follows: 

\begin{align}
    & X' = \textsc{Conv1D}(X),\nonumber \\
    & (g_1, g_2, \ldots, g_L) = \textsc{METIS}(G) \quad \text{where} \quad g_i = (V_i, E_i), \nonumber \\
    & s_i = \textsc{AveragePooling}(X'_{V_i}) \quad \text{for each} \quad g_i ,\nonumber \\
    & S = \textsc{Flatten}([s_1, s_2, \ldots, s_L]) .\nonumber
\end{align}

\noindent Thus, $S$ is a sequence of patches that can be fed into the transformer model for further processing.

\subsection{Sequential Modeling with Transformers}

After converting different flow data into the sequential structure, we leverage a transformer model to capture the spatio-temporal dependencies. The transformer model consists of a stack of $N$ identical layers, each containing a multi-head self-attention mechanism and a feed-forward neural network. The multi-head self-attention mechanism allows the model to focus on fine-grained spatio-temporal correlations and dynamics within the input sequence.

\textbf{Implementation details:} For spatio-temporal flow prediction, we adopt an encoder-decoder architecture for the transformer model.
The overall flow data $S$ includes both historical records $S_h$ and prediction horizons. 
The encoder masks the future data and processes only the historical data $S_h$. After obtaining the encoder output $Z$, we concatenate the embedding $Z$ with the mask token embedding $E_m$ to construct the complete embedding map.  The decoder then performs self-attention over the entire sequence of embeddings. The process is formulated as follows:
\begin{align}
    & S_h = \textsc{Mask}(S), \\
    & \text{Encoder:} \quad Z_e = \textsc{Encoder}(S_h), \nonumber \\
    & Z_d = \textsc{Concat}(Z_e, E_m), \\
    & \text{Decoder:} \quad Y = \textsc{Decoder}(Z_d), \nonumber
\end{align}

\noindent where $Z$ is the output of the encoder, which is then used as input to the decoder. $Y$ is the output of the decoder, which represents the predicted results.

\subsection{Spatio-Temporal Memory Retrieval Augmentation}

\begin{figure*}[t!]
    \centering
    \includegraphics[width=0.8\linewidth]{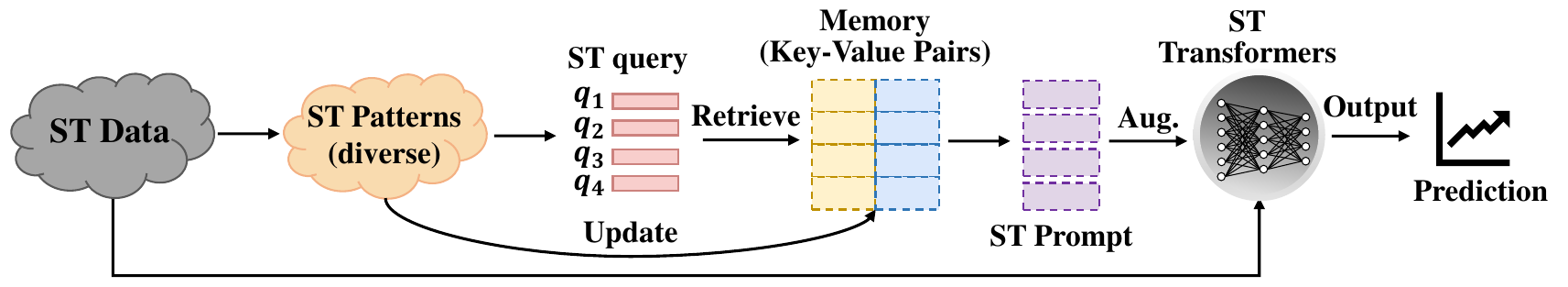}
    \caption{ST-MRA workflow in UniFlow.}
    \label{fig:MRA}
\end{figure*}

We design Spatio-Temporal Memory Retrieval Augmentation (ST-MRA) to memorize and utilize useful spatio-temporal patterns.
ST-MRA aims to enhance prediction accuracy by enabling cross-learning across different types of flow data. It does so by retrieving similar prompts from shared memories, thereby informing and improving current predictions. This approach addresses the challenge of integrating diverse spatio-temporal patterns, ultimately leading to more robust and informed predictions.
We demonstrate the workflow of ST-MRA in Figure~\ref{fig:MRA}.  The main steps of ST-MRA are as follows:

\begin{enumerate}[leftmargin=*]
\item Formulate a spatio-temporal query based on the input data;
\item Retrieve relevant memories to generate useful prompts;
\item Utilize the retrieved prompts to augment the input data.
\end{enumerate}

\noindent In the following subsections, we elaborate on the details of our proposed ST-MRA.

\subsubsection{\textbf{Spatio-Temporal Query Formulation}}
We extract spatio-temporal patterns from the input data to create queries, considering both temporal and spatial perspectives. 
Specifically, we consider two perspectives to obtain queries to represent patterns of the input flow data: temporal and spatial perspectives. Temporal queries are designed to capture patterns and trends over time, ensuring that the model understands temporal dynamics. In contrast, spatial queries focus on depicting spatial correlations, allowing the model to recognize and leverage the relationships between different locations. This dual approach ensures a comprehensive representation of spatio-temporal flow data.

\textbf{Temporal queries.}
To effectively understand temporal dynamics, we analyze data through two key characteristics: the time domain and the frequency domain. This enables a comprehensive understanding of both short-term and long-term patterns.

\begin{itemize}[leftmargin=*]
    \item \textit{Time domains.} This refers to the direct representation of data over time, capturing the sequence of changes. Urban phenomena often depend on consecutive time steps, such as traffic flow being influenced by the previous minutes or hours. 
    \item \textit{Frequency domains.} This domain uses Fourier analysis to transform temporal data into frequency components, revealing periodic and cyclical behaviors that are not visible in the time domain. Spatio-temporal dynamics often show periodic patterns, such as daily and weekly traffic flows.
\end{itemize}

Each of these domains offers unique perspectives and advantages that are essential for comprehensively modeling temporal patterns. Specifically, we use self-attention to obtain time-domain patterns and Fast Fourier Transform (FFT) to obtain frequency-domain patterns. The process can be formulated as follows: 

\begin{align}
    & E_t = \textsc{SelfAttention}(S_h), \nonumber \\
    & E_f = \textsc{FFT}(S_h), \nonumber
\end{align}

\noindent where $E_t$ and $E_f$ are the extracted time-domain and frequency-domain patterns, respectively.

\textbf{Spatial queries.}
Spatio-temporal data differ from traditional time series data due to their inherent spatial correlations, making effective modeling of spatial patterns crucial for accurate predictions. 
Both grid-based and graph-based data have unique spatial organizations but share the common concept that their spatial organizations both represent relationships between different parts of a city. 
Thus, we design an adaptive approach to extract spatial patterns for both types of data. Based on the learned patterns in both the time and frequency domains, we adaptively determine the spatial topology between different locations. This adaptive learning process is formulated as follows:

\begin{align}
    & A_t = softmax(\textsc{ReLU}(E_t\cdot E_t^T)), \nonumber \\
    & A_f = softmax(\textsc{ReLU}(E_f\cdot E_f^T)), \nonumber
\end{align}

\noindent where $A_t$ and $A_f$ are the learned spatial topology based on time-domain patterns and frequency-domain patterns, respectively.

Using the learned graph topology, we employ a Graph Convolutional Network (GCN) to learn the spatial representations between different locations:

\begin{equation}
    E_{st} = \textsc{GCN}(E_t, A_t), \ \ 
    E_{sf} = \textsc{GCN}(E_f, A_f), 
\end{equation}

\noindent where $E_{st}$ and $E_{sf}$  are extracted spatial patterns, i.e., spatial queries. It is worth noting that these spatial queries are not purely from a spatial perspective but incorporate entangled spatio-temporal relationships.

\subsubsection{\textbf{Memory Retrieval Augmentation}}

As shown in Figure~\ref{fig:MRA}, we design shared spatio-temporal memories to store and recall spatio-temporal patterns. These memories act as powerful repositories for retrieving relevant pattern, enabling the model to access and leverage already-learned patterns effectively.

\textbf{Spatio-temporal memories.}
We maintain external memories for spatio-temporal retrieval augmentation. Based on well-established spatio-temporal knowledge~\cite{zheng2014urban}, we categorized these memories into four groups:

\begin{itemize}[leftmargin=*]
    \item \textit{Time-domain memory:} Stores time-domain patterns extracted from historical data, capturing sequential dependencies.
    \item \textit{Frequency-domain memory:} Stores frequency-domain patterns, capturing periodic and cyclical behaviors.
    \item \textit{Time-spatial memory:} Stores spatial patterns, capturing spatial correlations between different locations based on time-domain patterns.
    \item \textit{Frequency-spatial memory:} Stores spatial patterns, capturing spatial correlations and relationships between different locations based on frequency-domain patterns.
\end{itemize}

\textbf{Memory structure.}
Each of these memories is structured as key-value pairs $(K, V)$, where both keys and values are  $N \times D$ learnable embeddings. Here, $N$ represents the number of memory units in the base, which is a hyper-parameter, and $D$  is the embedding dimension, which equals the output size of the encoder.

\textbf{Prompt retrieval.}
These memories store knowledge from the time domain, frequency domain, and adaptive spatial topology derived from both domains.
The retrieval process involves querying these memories to obtain useful prompts. We use the learned temporal and spatial patterns as queries to guide this process. For each query, we calculate the similarity between the query and each key unit in the memory, then extract the corresponding value based on these similarity scores.
The process is formulated as follows:

\begin{align}
    & Q_i \in \{E_t, E_f, E_{st}, E_{sf}\},\  K_i \in \{K_t, K_f, K_{st}, K_{sf}\} ,\nonumber \\
    & \alpha_i = softmax(Q_i \cdot K_i^T), \nonumber \\
    & P_i = \sum_{j} \alpha_{ij}V_i, \text{ where } V_i \in \{V_t, V_f, V_{st}, V_{sf}\},\nonumber
\end{align}

\noindent where $Q_i$ represents the input queries, $K_i$ represents the keys in the memory, and $V_i$ represents the corresponding values.

\textbf{Augmentation.} The augmentation process is straightforward. After obtaining the retrieved prompt $P_i$, we add it to the input embedding of the decoder for augmentation and feed it into the transformer model for further processing.  This augmented input is then passed through the decoder to generate the final predictions.

\begin{table*}[t!]
\caption{Results of short-term prediction on six datasets in terms of MAE and RMSE. We use the average prediction errors over all prediction steps. Bold denotes the best results and \underline{underline} denotes the second-best results.}
\label{tbl:short}
\begin{threeparttable}
\resizebox{2.0\columnwidth}{!}{
\begin{tabular}{ccccccccccccc}
\toprule
& \multicolumn{8}{c}{\textbf{Grid Data}} & \multicolumn{4}{c}{\textbf{Graph Data}}  \\
\cmidrule(lr){2-9} \cmidrule(lr){10-13}
& \multicolumn{2}{c}{\textbf{TaxiBJ}} & \multicolumn{2}{c}{\textbf{FlowSH}} & \multicolumn{2}{c}{\textbf{CrowdNJ}} & \multicolumn{2}{c}{\textbf{TaxiNYC}} & \multicolumn{2}{c}{\textbf{TrafficBJ}} & \multicolumn{2}{c}{\textbf{TrafficSH}} \\
\cmidrule(lr){2-3} \cmidrule(lr){4-5} \cmidrule(lr){6-7} \cmidrule(lr){8-9} \cmidrule(lr){10-11} \cmidrule(lr){12-13}
\textbf{Model} & \textbf{RMSE} & \textbf{MAE} & \textbf{RMSE} & \textbf{MAE} & \textbf{RMSE} & \textbf{MAE} & \textbf{RMSE} & \textbf{MAE} & \textbf{RMSE} & \textbf{MAE} & \textbf{RMSE} & \textbf{MAE} \\
\cmidrule(lr){1-1} \cmidrule(lr){2-3} \cmidrule(lr){4-5} \cmidrule(lr){6-7} \cmidrule(lr){8-9} \cmidrule(lr){10-11} \cmidrule(lr){12-13}
HA & 91.55 & 53.036 & 38.920 & 13.435 & 0.932 & 0.487 & 77.10 & 26.49 & 2.205 & 1.457 & 2.381 & 1.554 \\
ARIMA & 291 & 57.5 & 26.7031 & 9.152 & 0.989 & 0.4435 & 99.220 & 23.911 & 2.608 & 1.642 & 1.7496 & 0.620  \\
\cmidrule(lr){1-1} \cmidrule(lr){2-3} \cmidrule(lr){4-5} \cmidrule(lr){6-7} \cmidrule(lr){8-9} \cmidrule(lr){10-11} \cmidrule(lr){12-13}
STResNet & 37.96 & 26.55 & 59.82 & 45.63 & 0.718 & 0.511 & 26.88 & 14.81 & - &  - & - & - \\
ACFM & 30.95 & 19.87 & 46.92 & 24.95 & 0.468 & 0.284 & 20.82 & 9.85 & - &  - & - & - \\
STNorm & 31.21 & 19.00 & 28.46 & 11.88 & 0.384 & 0.231 & 26.94 & 10.43 & - &  - & - & - \\
STGSP & 27.31 & 17.54 & 38.77 & 17.54 & 0.410 & 0.263  & 25.94 & 10.52& - &  - & - & - \\
MC-STL & 38.50 & 28.51 & 46.06 & 33.83 & 0.727 & 0.504 & 36.75 & 26.01 & - &  - & - & - \\
MAU & 71.07 & 46.37 & 45.04 & 21.38 & 0.648 & 0.402 & 49.15 & 21.79 & - &  - & - & - \\
MIM & 68.18 & 42.40 & 47.29 & 22.49 & 0.715 & 0.399 & 24.53 & 9.151 & - &  - & - & - \\
SimVP & 35.58 & 21.67& 28.59 & 15.87 & 0.282 & 0.191 & 19.69 & 9.08 & - &  - & - & - \\
TAU & 26.43 & 15.86 & 26.04 & 15.22 & 0.326 & 0.219 & 19.46 & 9.08& - &  - & - & - \\
PromptST & 27.42 & 16.12 & 23.01 & 9.37 & 0.306 & 0.161 & 22.82 & 8.24 & - &  - & - & - \\
UniST &  \underline{23.67} & \underline{14.04} & \underline{19.95} & \underline{9.10} & \underline{0.191} & \underline{0.119} & \underline{17.55} & \underline{5.85} & -&- & -&- \\
\cmidrule(lr){1-1} \cmidrule(lr){2-3} \cmidrule(lr){4-5} \cmidrule(lr){6-7} \cmidrule(lr){8-9} \cmidrule(lr){10-11} \cmidrule(lr){12-13}
STGCN & -&- &-& -&- &- &- &- &2.53 &1.94 &2.20 &1.66 \\
DCRNN & -&- &- &- &- &- &- &- & 2.05 & 1.47 & 2.09 & 1.49 \\
GWN& -&- &- &- &- &- &- &- & 2.40 & 1.82 & 2.15 & 1.56 \\
MTGNN & -&- &- &- &- &- &- &- & 1.76 & 1.23 & \underline{1.85} & \underline{1.27} \\
AGCRN  & -&- &- &- &- &- &- &- &  2.37 & 1.78 & 2.55 & 1.91 \\
GTS & -&- &- &- &- &- &- &- & 2.44 & 1.89 & 2.84 & 2.21 \\
STEP & -&- &- &- &- &- &- &- & 2.11 & 1.56 & 2.15 & 1.57 \\
\cmidrule(lr){1-1} \cmidrule(lr){2-3} \cmidrule(lr){4-5} \cmidrule(lr){6-7} \cmidrule(lr){8-9} \cmidrule(lr){10-11} \cmidrule(lr){12-13}
STID & 25.55 & 16.36 & 21.19 & 12.92 & 0.234 & 0.160 & 18.49 & 8.32 & \underline{1.75} & \underline{1.16} & 2.06 & 1.39\\
PatchTST & 53.36 & 30.55 & 28.17 & 10.69 & 0.465 & 0.223& 50.45& 17.03&2.06 & 1.368 & 2.24 & 1.47 \\
PatchTST-v2 & 60.55 & 33.62 & 31.79 & 12.16 & 0.811 & 0.403& 58.61 & 21.27 & 2.53 & 1.67 & 2.78 & 1.82 \\
iTransformer & 42.17 & 24.05 & 25.91 & 10.19 & 0.466 & 0.216 & 45.19 & 45.19 &  2.04 & 1.35 & 2.29 & 1.52\\
Time-LLM & 51.20 & 29.55 & 28.19 &  10.57 & 0.405 &0.210 & 52.94 & 17.65 & 2.07 & 1.37 & 2.23 & 1.47\\

\cmidrule(lr){1-1} \cmidrule(lr){2-3} \cmidrule(lr){4-5} \cmidrule(lr){6-7} \cmidrule(lr){8-9} \cmidrule(lr){10-11} \cmidrule(lr){12-13}

\textbf{UniFlow} & \textbf{20.35} & \textbf{12.14} & \textbf{15.18} & \textbf{6.08} & \textbf{0.180} & \textbf{0.107} & \textbf{16.02} & \textbf{5.81} & \textbf{1.72} & \textbf{1.12}  & \textbf{1.84} & \textbf{1.24} \\

\bottomrule

\end{tabular}}
\end{threeparttable}
\end{table*}





\section{Performance Evaluations}
 experiments to evaluate UniFlow's performance. 

\subsection{Experimental Settings}

\subsubsection*{\textbf{Datasets}}
We evaluate UniFlow using nine spatio-temporal datasets\footnote{We select cities with both grid-based and graph-based datasets.}, encompassing both grid-based and graph-based structures. The grid-based datasets include TaxiBJ, TaxiNYC, CrowdNJ, CrowdBJ, FlowSH, and PopSH. The graph-based datasets consist of TrafficBJ, TrafficSH, and TrafficNJ, where the road networks within the cities form the graph topology. It is noteworthy that the graph datasets we utilize are large-scale, containing over 10,000 nodes, in contrast to the smaller spatio-temporal graph datasets that were commonly used in the past~\cite{shao2022spatial,li2023generative}. Detailed descriptions and characteristics of each dataset are provided in Appendix Table~\ref{tbl:append_data_grid} and Table~\ref{tbl:append_data_graph}. For each dataset, we partition the data into training, validation, and testing sets with a ratio of 6:2:2.

\subsubsection*{\textbf{Baselines}} We compare UniFlow with a broad collection of state-of-the-art spatio-temporal prediction models, which can be categorized into four main groups\footnote{We provide detailed description of selected baselines in Appendix~\ref{sup:baseline}. }:

\begin{itemize}[leftmargin=*]
    \item \textit{Classical methods.} These include methods such as HA and ARIMA, which use predefined rules or simple mathematical models to make predictions without requiring extensive training.
    \item \textit{Deep spatio-temporal prediction models (grid):}  These  include STResNet~\cite{zhang2017deep}, ACFM~\cite{liu2018attentive}, STNorm~\cite{deng2021st}, STGSP~\cite{zhao2022st}, MC-STL~\cite{zhang2023mask}, MAU~\cite{chang2021mau}, MIM~\cite{wang2019memory}, SimVP~\cite{gao2022simvp}, TAU~\cite{tan2023temporal}, STUD~\cite{shao2022spatial}, PromptST~\cite{zhang2023promptst}, UniST~\cite{yuan2024unist}. These deep learning-based models are designed to handle grid-based flow data, leveraging techniques that can capture spatial dependencies in a grid layout.
    \item \textit{Deep spatio-temporal prediction models (graph):} These  include STGCN~\cite{yu2018spatio}, GWN~\cite{wu2019graph}, MTGNN~\cite{wu2020connecting}, GTS~\cite{shang2021discrete}, DCRNN~\cite{li2018diffusion}, STEP~\cite{shao2022pre}, AGCRN~\cite{bai2020adaptive}. These methods focus on graph-based flow data, utilizing GNN-based techniques to model the complex relationships and interactions within graph structures.
    \item \textit{Time series prediction models:} These  include PatchTST~\cite{nie2022time}, iTransformer~\cite{liu2023itransformer}, Time-LLM~\cite{jin2023time}. These approaches are proposed for multivariate time series prediction. We also train a one-for-all PatchTST, denoted as PatchTST-v2, as the baseline.
\end{itemize}

\subsubsection*{\textbf{Experimental Configuration}} The model is configured with 4 encoder layers and 4 decoder layers. The hidden size is set to 256. The maximum number of training epochs is 200, with early stopping employed to prevent overfitting. All baseline models are also trained with a maximum of 200 epochs for a fair comparison. The batch size is adjusted for each dataset to ensure relatively similar training iterations across different datasets. Each memory contains 512 embeddings. The learning rate is initially set to $5\times 10^{-4}$ and decays to $5\times 10^{-5}$  after 200 epochs.

\subsection{Short-Term Prediction}

\subsubsection*{\textbf{Setups}}
For short-term prediction, we define both the input step and prediction horizon as 12, denoted as $12 \rightarrow 12$. For the baseline models, we train a specific model tailored to each dataset, while UniFlow is evaluated across all datasets. It is worth noting that the step size of 12 corresponds to different temporal scales for different datasets. For instance, in the Traffic dataset, 12 represents three hours, whereas in the TaxiBJ dataset, 12 corresponds to six hours.

\subsubsection*{\textbf{Results}}

\begin{table}[t!]
\caption{Results of long-term prediction on three datasets in terms of MAE and RMSE. }
\label{tbl:long}
\resizebox{1.0\columnwidth}{!}{
\begin{tabular}{ccccccc}
\toprule
& \multicolumn{4}{c}{\textbf{Grid Data}} & \multicolumn{2}{c}{\textbf{Graph Data}}  \\
\cmidrule(lr){2-5} \cmidrule(lr){6-7}
& \multicolumn{2}{c}{\textbf{TaxiBJ}} & \multicolumn{2}{c}{\textbf{FlowSH}} & \multicolumn{2}{c}{\textbf{TrafficBJ}} \\
\cmidrule(lr){2-3} \cmidrule(lr){4-5} \cmidrule(lr){6-7}
\textbf{Model} & \textbf{RMSE} & \textbf{MAE} & \textbf{RMSE} & \textbf{MAE} & \textbf{RMSE} & \textbf{MAE} \\
\cmidrule(lr){1-1} \cmidrule(lr){2-3} \cmidrule(lr){4-5} \cmidrule(lr){6-7}
HA & 74.16 &  43.55 & 41.84 & 18.02 & 2.20 & 1.50 \\
ARIMA & 101 & 55.69 & 53.01 & 21.73 & 2.75 & 1.87\\
\cmidrule(lr){1-1} \cmidrule(lr){2-3} \cmidrule(lr){4-5} \cmidrule(lr){6-7}
STResNet & 43.98 & 29.49 & 72.87 & 50.88 & -&- \\
ACFM & 59.85 & 40.50 & 45.92 & 27.45 & -&- \\
STNorm & 41.69 & 25.26 & 39.59 & 21.60 & -&- \\
STGSP & 26.92 & 17.09 & 33.86 & 19.37 &-&- \\
MC-STL & 48.06 & 35.25 & 50.61 & 36.62 &-&- \\
MAU & 37.32 & 21.59 & 38.86 & 17.98 & -&- \\
MIM & 36.52 & 20.92 & 41.73 & 22.92 & -&- \\
SimVP & 33.78 & 20.13 & 31.04 & 16.74 & -&- \\
TAU &31.28 & 18.59 & 34.68 & 20.58 &-&- \\
PromptST & 38.83 & 22.52 & 34.83 & 13.80 & -&- \\
UniST & \underline{25.21} & \underline{15.66} & 34.18 & 14.30 & - & -\\
\cmidrule(lr){1-1} \cmidrule(lr){2-3} \cmidrule(lr){4-5} \cmidrule(lr){6-7}
STGCN & -& -&- &-&3.15 &2.57 \\
DCRNN &-& -&- &- & 2.79 & 2.22\\
GWN &-& -&- &- & 3.12 & 2.52 \\
MTGNN & -& -&- &- & 2.93 & 2.30\\
AGCRN & -& -&- &- & 2.23 & 1.64\\
GTS & -& -&- &-& 3.25 & 2.66 \\
STEP & -& -&- &- & 2.44 & 1.81 \\
\cmidrule(lr){1-1} \cmidrule(lr){2-3} \cmidrule(lr){4-5} \cmidrule(lr){6-7}
STID & 29.69 & 19.46 & 29.51 & 17.99 & \underline{1.89} & \underline{1.28} \\
PatchTST & 34.43 & 18.56 & 26.41 & \underline{9.50} & 1.91 & 1.22 \\
iTransformer & 36.62 & 19.76 & \underline{25.54} & 10.35 & 1.93 & 1.27 \\
Time-LLM & 33.85 & 18.64 & 44.69 & 16.52 & 1.92 & 1.25 \\
\cmidrule(lr){1-1} \cmidrule(lr){2-3} \cmidrule(lr){4-5} \cmidrule(lr){6-7}
\textbf{UniFlow} & \textbf{23.63} & \textbf{13.75} & \textbf{19.10} & \textbf{7.65} & \textbf{1.81} & \textbf{1.19} \\
\bottomrule
\end{tabular}}
\end{table}

Table~\ref{tbl:short} illustrates the results of short-term prediction.\footnote{The remaining results can be found in Appendix~\ref{sup:result}.} As shown, video prediction models, such as MAU, MIM and SimVP, did not perform very well. In contrast, urban spatio-temporal baselines, such as STID, STResNet, ACFM, PromptST, and UniST, achieve better performance than video models. This indicates that the spatio-temporal dynamics in urban flow data are quite different from those in video data, requiring specialized designs for flow data. For graph-based data, MTGNN exhibits relatively good performance. 
These results differ from previous works~\cite{shao2022pre, shao2022spatial}, likely due to the very large graph topology in our dataset, with more than 10,000 nodes.

Multivariate time series approaches also fail to perform well. The one-for-all PatchTST performed even worse than separate PatchTST models, indicating its inability to learn mutual benefits across different datasets. Time-LLM performed poorly in urban spatio-temporal flow prediction, indicating the necessity of training a foundation model specifically for urban spatio-temporal flow data.

UniFlow achieves the best performance across all datasets, including both grid-based and graph-based data.  Compared to the best baseline, UniFlow with single model shows a relative improvement of 9.1\%.  The superiority of UniFlow highlights the potential to unify different flow data for enhancing emergent capabilities.

\subsection{Long-Term Prediction}

\subsubsection*{\textbf{Setups}}
For long-term prediction, we extend our evaluation to capture the model's ability to predict over extended periods. In this setup, both the input step and prediction horizon are set to 64, denoted as $64 \rightarrow 64$.  This evaluation aims to test the model's capability to  maintain accuracy over longer prediction horizons.

\subsubsection*{\textbf{Results}} 

Table~\ref{tbl:long} presents the results of long-term prediction. As the prediction horizon increases, we observe that video prediction models generally show improvement, while urban spatio-temporal models, including both grid-based and graph-based approaches, exhibit a decline in performance. This suggests that video models are better suited to handling long-term scenarios. Consistent with its superior performance in short-term prediction, UniFlow also achieves the best performance in the long-term prediction setting. Compared with the best baseline, it achieves a relative improvement of 11.9\% in terms of RMSE.

\begin{figure}[t!]
    \centering
    \includegraphics[width=\linewidth]{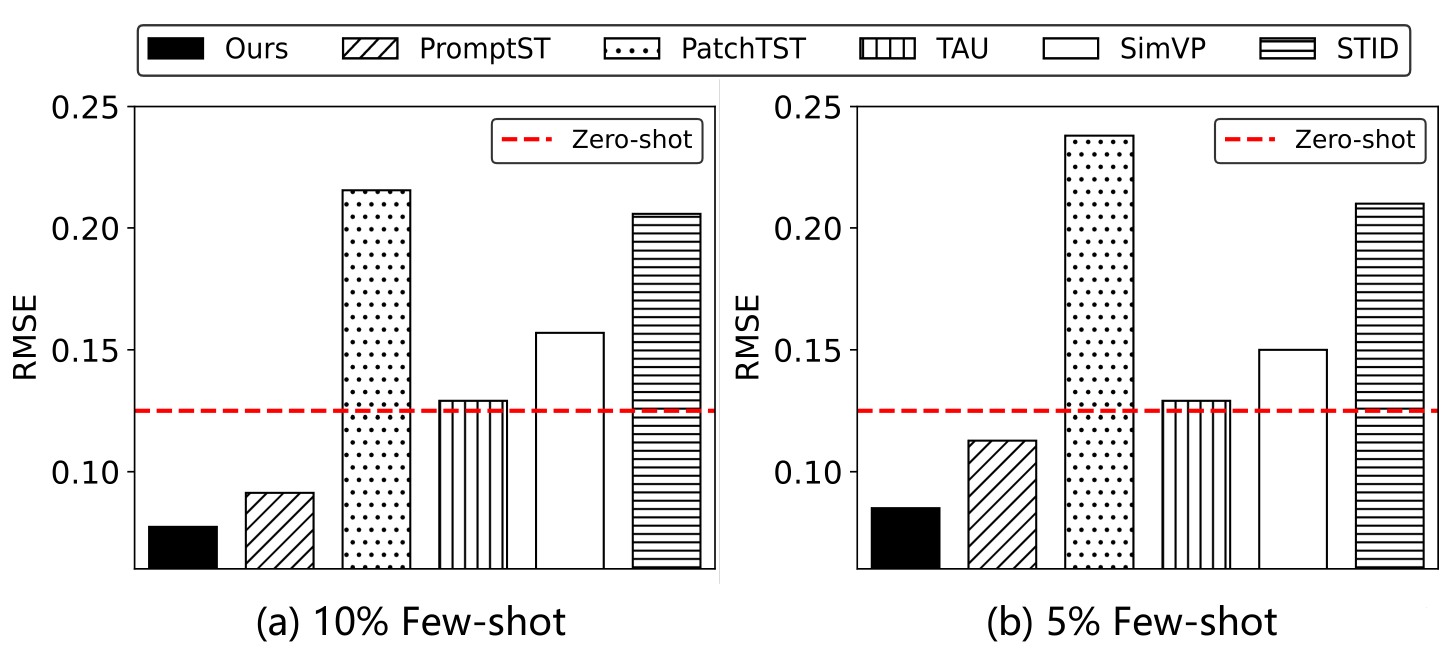}
    \caption{Performance comparison of few-shot predictions between UniFlow and baseline models \textcolor{black}{on the CrowdSH dataset}. The red dashed line represents zero-shot performance. }
    \label{fig:few_zero}
\end{figure}

\subsection{Few-Shot and Zero-Shot Capabilities}

\subsubsection*{\textbf{Setups}} 

Foundation models have demonstrated remarkable generalization abilities, performing well even with limited data or no data at all. To verify whether our proposed model inherits this capability, we design experiments for both few-shot and zero-shot predictions. \textcolor{black}{We divide the datasets into two groups: source datasets and target datasets, with the CrowdSH dataset designated as the target and the remaining datasets as the source.} The model is trained solely on the source datasets.

For few-shot prediction, we fine-tune the model with a limited amount of data from the target datasets and evaluate its performance on the same target datasets. This setup challenges the model to learn effectively from sparse data and aims to assess its ability to generalize from limited information. We select small portions of the target dataset for fine-tuning, specifically 5\% and 10\%.

For zero-shot prediction, we do not provide any training data from the target domain. Instead, we directly evaluate the model's performance on the target datasets without any fine-tuning.

\subsubsection*{\textbf{Results}} 
Figure~\ref{fig:few_zero} shows the comparison results of UniFlow against selected baselines with relatively good performance in short-term prediction. As we can observe, with limited training data, UniFlow exhibits the best few-shot learning capabilities, nearly achieving results comparable to normal prediction with complete training data. However, baseline approaches show significant performance degradation, indicating their limitations in sparse data scenarios. The red dashed line denotes the zero-shot performance of UniFlow, which surprisingly surpasses most baseline approaches that have some training data, demonstrating its superior generalization ability.

\begin{figure}[t!]
    \centering
    \includegraphics[width=0.75\linewidth]{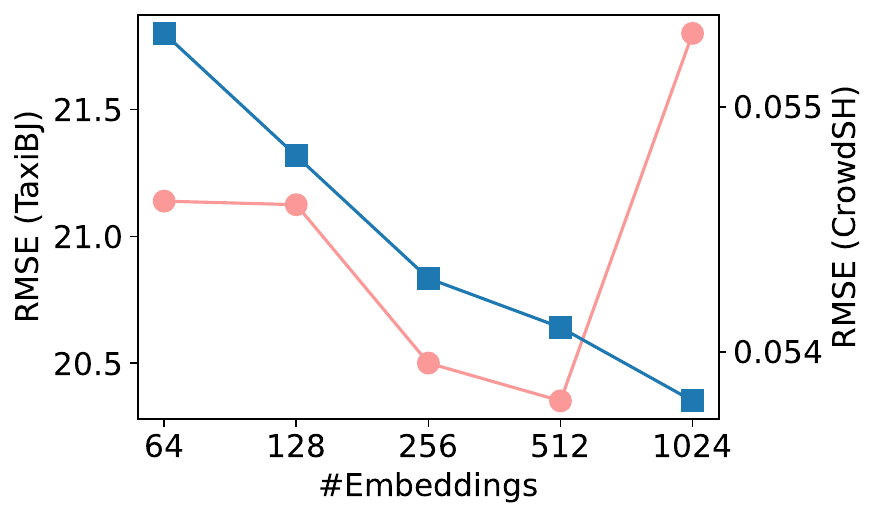}
    \caption{Results of ablation studies of UniFlow in terms of the number of units in each memory. }
    \label{fig:ablation_num}
\end{figure}

\begin{figure}[t!]
    \centering
    \includegraphics[width=\linewidth]{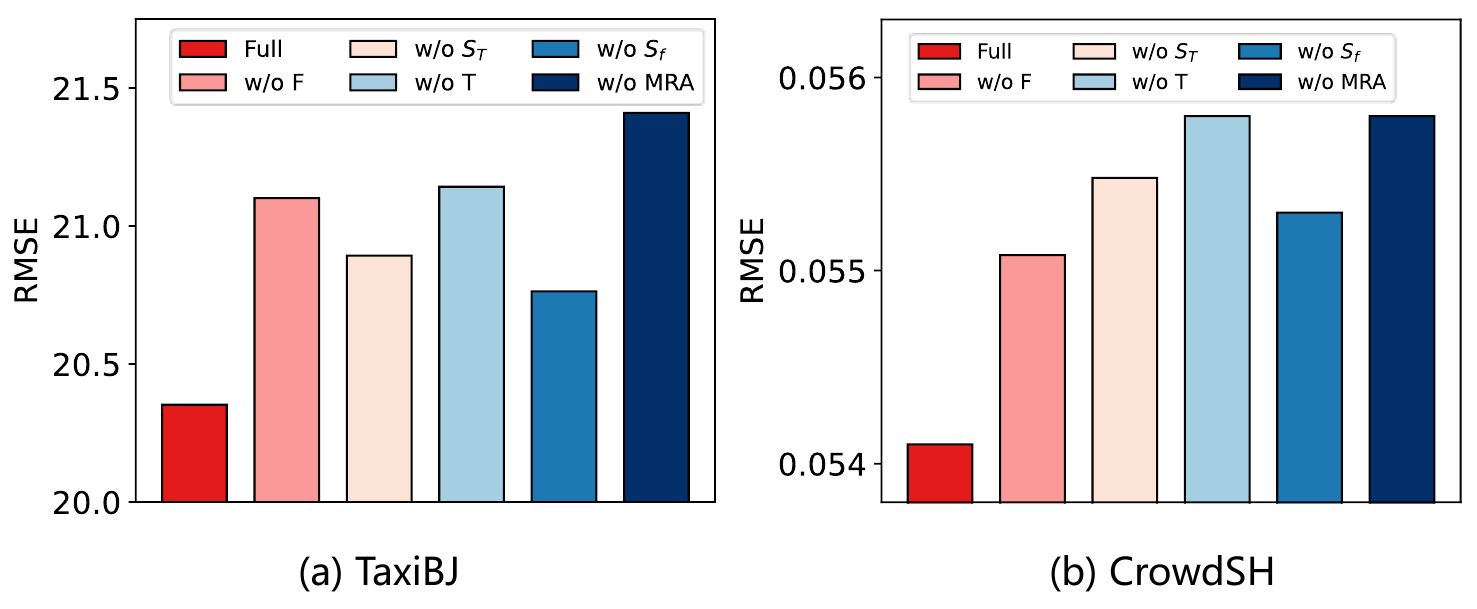}
    \caption{\textcolor{black}{Results of ablation studies of UniFlow in terms of the type of memory.}}
    \label{fig:ablation_type}
\end{figure}

\subsection{Ablation Study}

We perform ablation studies to examine how the different designs in ST-MRA contribute to the final performance. Our ablation studies focus on two perspectives. The first is the number of memory units (embeddings) in each memory. The second is the type of memories.

\subsubsection*{\textbf{Number of Memory Units}} Figure~\ref{fig:ablation_num} shows the results with different numbers of memory units. As we can observe, increasing the number of units initially reduces the prediction RMSE, indicating that memories with higher representation capabilities benefit the prompt retrieval process. However, beyond a certain point, further increasing the number of units leads to an increase in RMSE. This can be attributed to the fact that the memory stores prototypes representing typical spatio-temporal patterns. When the number of prototypes becomes too large, it becomes difficult to learn and categorize these typical prototypes. As a result, the prompt retrieval process becomes less effective.

\subsubsection*{\textbf{Type of ST-MRA}}  Figure~\ref{fig:ablation_type} shows the results of UniFlow and its degraded versions by removing certain parts of the ST-MRA. 
\textcolor{black}{We use the notations $T$, $F$, $S_T$, and $S_F$ to represent time-domain, frequency-domain, time-derived spatial, and frequency-derived spatial prompts, respectively. Additionally, we include an ablation study with all memory units removed, indicated as "w/o MRA".}
The results show that removing any memory significantly degrades the prediction performance. This indicates that they all contribute uniquely and critically to the overall model's effectiveness.

\begin{table}[t!]
\caption{Short-term prediction performance with different noise levels on TrafficBJ (graph) in terms of RMSE and MAE. We use the average prediction errors over all prediction steps. }
\label{tbl:graph-noise-short}
\begin{threeparttable}
\resizebox{1.0\columnwidth}{!}{
\begin{tabular}{ccccccc}
\toprule
\multirow{2}{*}{\textbf{Model}}&\multicolumn{2}{c}{\textbf{1\%}} & \multicolumn{2}{c}{\textbf{5\%}} & \multicolumn{2}{c}{\textbf{10\%}} \\
\cmidrule(lr){2-3} \cmidrule(lr){4-5} \cmidrule(lr){6-7} 
& \textbf{RMSE} & \textbf{MAE} & \textbf{RMSE} & \textbf{MAE} & \textbf{RMSE} & \textbf{MAE} \\
\cmidrule(lr){1-1}\cmidrule(lr){2-3} \cmidrule(lr){4-5} \cmidrule(lr){6-7} 
STGCN		&2.497	&1.972	&3.292	&2.662	&4.254	&2.965\\
DCRNN		&2.119	&1.542	&2.947	&2.277	&3.212	&2.519\\
GWN	&	2.448	&1.867	&2.935	&2.307	&3.149	&2.497\\
MTGNN		&1.776	&1.253	&2.214	&1.661	&2.701	&2.070\\
AGCRN		&2.408	&1.825	&2.942	&2.300	&3.326	&2.616\\
GTS&	2.447	&1.901	&2.624	&2.066	&3.003	&2.380\\
STEP	&2.119	&1.570	&2.401	&1.834	&2.927	&2.267\\
PatchTST & 2.09& 1.38 & 2.32 & 1.63 & 2.79 & 2.11 \\
iTransformer & 2.05 & 1.35 & 2.13 & 1.47 & 2.37 & 1.73\\
Time-LLM& 2.08 & 1.37 & 2.08 & 1.37 & 2.10 & 1.39 \\

\cmidrule(lr){1-1}\cmidrule(lr){2-3} \cmidrule(lr){4-5} \cmidrule(lr){6-7} 
\textbf{UniFlow} & 1.75 &  1.15 & 1.76 & 1.15 & 1.76 & 1.16  \\
\bottomrule
\end{tabular}}
\end{threeparttable}
\end{table}

\begin{table}[t!]
\caption{Short-term prediction performance with different noise levels on  TaxiBJ (grid) in terms of RMSE and MAE. We use the average prediction errors over all prediction steps. }
\label{tbl:grid-noise-short}
\begin{threeparttable}
\resizebox{1.0\columnwidth}{!}{
\begin{tabular}{ccccccc}
\toprule
\multirow{2}{*}{\textbf{Model}}&  \multicolumn{2}{c}{\textbf{1\%}} & \multicolumn{2}{c}{\textbf{5\%}} & \multicolumn{2}{c}{\textbf{10\%}} \\
\cmidrule(lr){2-3} \cmidrule(lr){4-5} \cmidrule(lr){6-7} 
& \textbf{RMSE} & \textbf{MAE} & \textbf{RMSE} & \textbf{MA} & \textbf{RMSE} & \textbf{MAE}  \\
\cmidrule(lr){1-1}\cmidrule(lr){2-3} \cmidrule(lr){4-5} \cmidrule(lr){6-7} 
STResNet&  39.38&27.78&71.70&48.85&139.8&92.1\\
ACFM &  30.90 & 19.89 &  31.388 &  20.28 & 32.66 & 21.24 \\
STNorm& 31.36&19.16&34.48&21.99&42.18&28.09\\
STGCP&  27.31&17.54&27.31&17.54&27.31&17.54\\
MC-STL&  38.53&28.54 &38.58 &28.57 &38.71&28.64\\
MAU&  71.07 &46.38 & 71.05& 46.37 & 71.02 & 46.35\\
MIM&  76.29&48.01& 76.31&48.02 &76.38 & 48.07\\
SimVP& 35.61 &21.70 &36.87 &23.12 &40.51 & 26.79\\
TAU& 26.59&15.96&27.80&17.06&31.27&19.91\\
PromptST&  27.42& 16.12&27.43&16.14& 27.45& 16.16\\
UniST& 23.87 & 14.59& 23.89& 14.60& 23.90& 14.61\\
STID &  25.59 & 16.41 & 26.48 & 17.48 & 29.09 & 20.22\\
PatchTST  &  55.37 & 33.20 & 86.38 & 61.98 & 137.5 & 105.7\\
iTransformer &  44.33 & 26.59 & 60.69 & 45.39 & 102.0 & 80.6\\

Time-LLM &33.79 &18.79 &33.99 &20.42 &35.25 &23.58\\
\cmidrule(lr){1-1}\cmidrule(lr){2-3} \cmidrule(lr){4-5} \cmidrule(lr){6-7} 
\textbf{UniFlow} &  20.83 & 12.56 & 22.76 & 14.44 & 22.97 & 14.52 \\
\bottomrule
\end{tabular}}
\end{threeparttable}
\end{table}

\subsection{Model Robustness}

To examine the robustness of our proposed model, we evaluate its performance under noise perturbations. Specifically, we introduce Gaussian noise into the spatio-temporal data, with a mean value of 0. We test three levels of noise standard deviation: 1\%,  5\%, and 10\% of the mean value of the ST data. By varying the noise levels, we can assess the model's robustness and its ability to handle different degrees of data corruption effectively.  
Table~\ref{tbl:graph-noise-short} and Table~\ref{tbl:grid-noise-short} illustrate the performance of our model and baseline methods under noise perturbations.  

 As we can observe from Table~\ref{tbl:append_data_graph}, models such as STGCN, DCRNN, MTGNN, AGCRN, GTS, and STEP experienced significant performance degradation under noise perturbations. Similarly, multivariate time series prediction approaches, including PatchTST, iTransformer, and Time-LLM, also exhibited substantial performance reductions. In contrast, our proposed model, UniFlow, maintained nearly steady performance under noise perturbations, demonstrating its robustness.
 Table~\ref{tbl:append_data_grid} corroborates these findings, showing that UniFlow maintains good performance even under noise perturbations, while the baseline models experience significant degradation. Additionally, UniST also achieves robust performance under noise perturbations, highlighting the advantage of robustness that a universal model can provide.

To further elaborate on the advantages of foundation models, it is important to highlight their capability to achieve outstanding performance and robustness through extensive training on vast amounts of data. Our proposed foundation model, UniFLow, leverages diverse and large-scale datasets to learn comprehensive spatio-temporal patterns. This allows UniST to generalize well across various tasks and handle diverse types of data perturbations effectively.

\subsection{Case Study}

To gain a deeper understanding of our designed ST-KRA process, we conducted case studies to investigate how different input spatio-temporal data interact with the memory. Specifically, for various spatio-temporal inputs, we calculated how each input retrieves information from the memory by recording the aggregation weights for each memory unit. We then compared the similarity between these weights using cosine similarity. This approach allowed us to investigate which scenarios exhibit similar behavior during the prompt retrieval process.

\begin{figure}[t!]
    \centering
    \includegraphics[width=\linewidth]{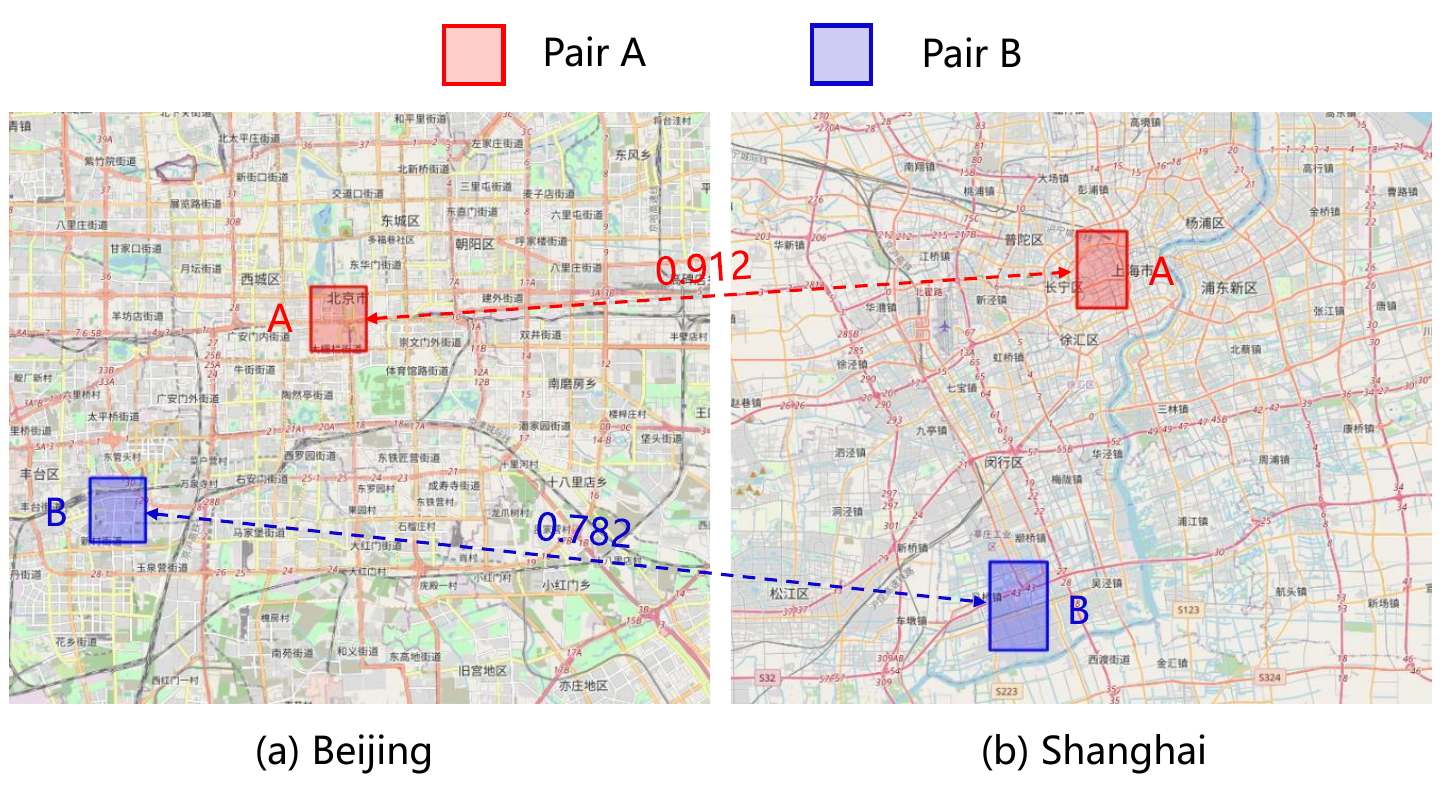}
    \caption{Case studies of the spatio-temporal memory retrieval augmentation. We select two pairs in Beijing and Shanghai that have similar retrieved prompts.}
    \label{fig:case}
\end{figure}

Figure~\ref{fig:case} illustrates two pairs of regions in Beijing and Shanghai, China. The numerical values above the dashed red and blue lines represent the similarity scores between the retrieval weights. The results are highly interpretable. Region A in Beijing encompasses central areas such as Changan Street and Qianmen Road, while Region A in Shanghai also represents the city's central part. Both regions exhibit high similarity scores, indicating similar spatio-temporal dynamics typically observed in urban centers. Region B in Beijing and Region B in Shanghai are both relatively remote residential areas. The similarity in prompt retrieval for these regions suggests that they share comparable spatio-temporal characteristics, likely due to similar patterns of residential activity and lower traffic density compared to the central regions.

\section{Conclusion}
In this paper, we develop a foundation model for unified urban spatio-temporal flow prediction. By utilizing the transformer architecture, we achieved uniform sequential modeling of both grid-based and graph-based flow data. We also design spatio-temporal memory retrieval augmentation  to generate useful prompts, which facilitate effective cross-learning of different data. Experimentally, the proposed one-for-all UniFlow model achieves state-of-the-art performance across various datasets. Further analysis reveals the efficacy of the memory retrieval augmentation. 
UniFlow demonstrates the potential to unify diverse types of spatio-temporal data to build a foundation model. We believe it inspires the development of more one-for-all foundation models in urban contexts.
In the future, we will explore additional enhancements, such as expanding the model's applicability to more complex spatio-temporal scenarios, such as meteorology and earth sciences.

\bibliographystyle{IEEEtran}
\bibliography{8.reference}

\clearpage


\appendix

\subsection{Datasets}\label{sup:dataset}

\begin{table*}[t!]
\caption{The basic information of grid-based spatio-temporal data.}
\label{tbl:append_data_grid}
\centering
\resizebox{1.9\columnwidth}{!}{
\begin{tabular}{cccccccc}
\toprule
Dataset & City & Type & Temporal Period & Spatial partition & Interval & Mean & Std \\
\hline
TaxiBJ & Beijing & Taxi flow&  2013/06/01 - 2013/10/30 & $32 \times 32$ & Half an hour & 97.543 & 122.174 \\
TaxiNYC & New York City & Taxi flow &  2015/01/01 - 2015/03/01 & $10 \times 20$ & Half an hour & 38.801 & 103.924 \\
FlowSH & Shanghai & Mobility flow &  2016/04/25 - 2016/05/01 & $20 \times 20$ & 15min & 31.935 & 137.926 \\
CrowdBJ & Beijing & Crowd flow &  2021/10/25 - 2021/11/21 & $28 \times 24$ & One hour & 0.367 & 0.411 \\
CrowdNJ & Nanjing & Crowd flow &  2021/02/02 - 2021/03/01 & $20 \times 28$ & One hour & 0.872 & 1.345 \\
PopSH & Shanghai & Dynamic population &  2014/08/01 - 2014/08/28 & $32 \times 28$ & One hour & 0.175 & 0.212 \\
\bottomrule
\end{tabular}}
\end{table*}

\begin{table*}[t!]
\caption{The basic information of Graph-based spatio-temporal data.}
\label{tbl:append_data_graph}
\centering
\resizebox{1.8\columnwidth}{!}{
\begin{tabular}{ccccccccc}
\toprule
Dataset & City & Type & Temporal Period & Interval & \#Nodes & \#Edges & Mean & Std \\
\hline

TrafficBJ & Beijing & Traffic speed & 2022/03/05 - 2022/04/05 & 15min& 13675& 24444& 6.837&  3.412\\
TrafficSH & Shanghai & Traffic speed & 2022/01/27 - 2022/02/27 & 15min & 21099& 39065& 7.815&  4.044\\
TrafficNJ & Nanjing & Traffic speed  & 2022/03/05 - 2022/04/05 & 15min & 13419& 25100& 6.699&  4.253\\

\bottomrule
\end{tabular}}
\end{table*}

\subsubsection{\textbf{Basic Information}}

We provide detailed information about the datasets used in our study, which fall into two categories: grid-based data and graph-based data. Table~\ref{tbl:append_data_grid} presents the basic information and statistics for the grid-based datasets, while Table~\ref{tbl:append_data_graph} offers similar details for the graph-based datasets. It is worth mentioning that we did not utilize popular spatio-temporal graph datasets such as PEMS-03, PEMS-07, and PEMS-Bay. The primary reason is the lack of corresponding grid-based spatio-temporal data for these cities. We have observed that graph-based data benefit more from grid-based data when both come from the same city.

\subsubsection{\textbf{Data Preprocessing}}

For each dataset, we divided it into three distinct segments: the initial 60\% was allocated for training, the subsequent 20\% for validation, and the remaining 20\% for testing. All datasets were normalized to a scale of $[0, 1]$. The prediction results presented are the denormalized values.

\subsection{Baselines}\label{sup:baseline}

\subsubsection{\textbf{Basic information}}

\begin{itemize}[leftmargin=*]
    \item \textbf{HA}: History Average uses the average of past data from corresponding periods to forecast future values.
    \item \textbf{ARIMA}: The Auto-Regressive Integrated Moving Average model is a widely-used statistical method for time series forecasting. It effectively analyzes and predicts time series data collected at regular intervals.
    \item \textbf{MIM}~\cite{wang2019memory}: This model leverages differential information between adjacent recurrent states to handle non-stationary properties. Multiple stacked MIM blocks enable the modeling of higher-order non-stationarity.
    \item \textbf{MAU}~\cite{chang2021mau}: The Motion-aware Unit broadens the temporal receptive fields of prediction units to capture inter-frame motion correlations. It includes an attention module and a fusion module for video prediction.
    \item \textbf{SimVP}~\cite{gao2022simvp}:A simple yet effective video prediction model built entirely on convolutional neural networks and using MSE loss, serving as a solid baseline in video prediction tasks.
    \item \textbf{TAU}~\cite{tan2023temporal}:  The Temporal Attention Unit decomposes temporal attention into intra-frame and inter-frame components, introducing differential divergence regularization to account for inter-frame variations.
    \item \textbf{STResNet}~\cite{zhang2017deep}: It utilizes residual neural networks to capture temporal closeness, periodicity, and trends in the data.
    \item \textbf{ACFM}~\cite{liu2018attentive}: Attentive Crowd Flow Machine model predicts crowd dynamics by employing an attention mechanism to adaptively aggregate sequential and periodic patterns.
    \item \textbf{STGSP}~\cite{zhao2022st}: This model emphasizes the importance of global and positional information in the temporal dimension for spatio-temporal prediction. It uses a semantic flow encoder for temporal positional signals and an attention mechanism for multi-scale temporal dependencies. 
    \item \textbf{MC-STL}~\cite{zhang2023mask}: It  employs mask-enhanced contrastive learning to effectively capture relationships in the spatio-temporal dimension.
    \item \textbf{STNorm}~\cite{deng2021st}: It proposes two types of normalization modules: spatial normalization and temporal normalization, which separately handle high-frequency and local components.
    \item \textbf{STID}~\cite{shao2022spatial}: This MLP-based spatio-temporal prediction model identifies indistinguishability in spatio-temporal dimensions, demonstrating efficiency and effectiveness in its design.
    \item \textbf{PromptST}~\cite{zhang2023promptst}: A state-of-the-art pre-training and prompt-tuning approach designed specifically for spatio-temporal prediction.
    \item \textbf{UniST}~\cite{yuan2024unist}: A universal model urban spatio-temporal prediction with grid-based data. It leverages diverse spatio-temporal masking strategies for pre-training and spatio-temporal knowledge-based prompt fine-tuning.
    \item \textbf{STGCN}~\cite{yu2018spatio}: Spatio-Temporal Graph Convolutional Network is a deep learning framework for traffic forecasting that leverages both spatial and temporal dependencies. It combines graph convolutional layers with convolutional sequence learning layers to model multi-scale traffic networks.
    \item \textbf{GWN}~\cite{wu2019graph}: Graph WaveNet is a CNN-based method designed to address the limitations of existing spatial-temporal graph modeling approaches. It proposed a self-adaptive adjacency matrix and stacked dilated causal convolutions to capture temporal dependencies effectively. 
    \item \textbf{MTGNN}~\cite{wu2020connecting}: MTGNN is a framework specifically designed for multivariate time series data. It automatically extracts uni-directed relations among variables through a graph learning module and integrates external knowledge like variable attributes.
    \item \textbf{GTS}~\cite{shang2021discrete}: GTS is a method that learns the structure of a graph simultaneously with a Graph Neural Network (GNN) for forecasting multiple time series. It parameterizes the graph structure through a neural network, enabling discrete graphs to be sampled differently, and optimizes the mean performance over the graph distribution. 
    \item \textbf{DCRNN}~\cite{li2018diffusion}: Diffusion Convolutional Recurrent Neural Network is a deep learning model for spatiotemporal forecasting. It models traffic flow as a diffusion process on a directed graph, capturing spatial dependencies through bidirectional random walks and temporal dependencies using an encoder-decoder architecture with scheduled sampling.
    \item \textbf{STEP}~\cite{shao2022pre}:Spatial-temporal Graph Neural Network Enhanced by Pre-training is a framework that incorporates a scalable time series pre-training model to improve multivariate time series forecasting. 
    \item \textbf{AGCRN}~\cite{bai2020adaptive}: AGCRN enhances Graph Convolutional Networks (GCNs) with two adaptive modules: Node Adaptive Parameter Learning (NAPL) and Data Adaptive Graph Generation (DAGG). It  captures fine-grained spatial and temporal correlations in traffic series without relying on pre-defined graphs.
    \item \textbf{PatchTST}~\cite{nie2022time}: It introduces patching and self-supervised learning for multivariate time series forecasting, segmenting the time series into patches to capture long-term correlations and processing different channels independently with a shared network.
    \item \textbf{iTransformer}~\cite{liu2023itransformer}:  This state-of-the-art model for multivariate time series employs attention and feed-forward operations on an inverted dimension, focusing on multivariate correlations.
    \item \textbf{Time-LLM}~\cite{jin2023time}: TIME-LLM is the state-of-the-art model to apply large language models to time series forecasting. It utilizes a reprogramming framework designed to repurpose LLMs for general time series forecasting, all while keeping the backbone language models intact.
\end{itemize}

\subsubsection{\textbf{Baseline implementation}}
\textbf{Dealing with very large graph.} Assuming the Graph-based ST data is \(X\in \mathbb{R}^{T\times N\times C}\), where $N$ corresponds to the total node number, $T$ represents the temporal period length and $C$ is the number of variables. The overly large N (over 10,000) leads to an excessively high computational complexity for training models. To better train baseline models, we evenly split the spatial graph into multiple subgraphs. We convert $X$ to \(X'\in \mathbb{R}^{T\times n\times m\times C}\) , where $m$ is the number of subgraphs and $n$ is the node number of each subgraph. When training the model, randomly select a subgraph from $m$ subgraphs as the training sample. The input of the model \({x_{in}}\in \mathbb{R}^{B\times h\times n\times C}\), and the target \({x_{tar}}\in \mathbb{R}^{B\times t\times n\times C}\). $B$ is batch size, $h$ is historical input steps, and $t$ is prediction horizons. The information of subgraphs for three datasets is shown in Table~\ref{tbl:node}.

\textbf{One-for-all models.} Among the selected baselines, PatchTST and UniST are channel-independent (variate-independent) models. To ensure a fair comparison, we train a one-for-all model for both.

\begin{table}
\caption{Information of subgraphs for different datasets}
\label{tbl:node}
\centering
\resizebox{0.6\columnwidth}{!}{
\begin{tabular}{cccc}
\toprule
Dataset & N & n & m  \\
\hline

TrafficBJ & 13675& 549& 25\\
TrafficSH & 21099& 541& 39\\
TrafficNJ & 13419& 497& 27\\
\bottomrule
\end{tabular}}

\end{table}

\subsection{Training and Evaluation}

\subsubsection{\textbf{Training}}

\begin{table}[t!]
\caption{Performance comparison of short-term prediction on the remaining datasets in terms of MAE and RMSE. We use the average prediction errors over all prediction steps. }
\label{tbl:app_short}
\begin{threeparttable}
\resizebox{1.0\columnwidth}{!}{
\begin{tabular}{ccccccc}
\toprule
& \multicolumn{4}{c}{\textbf{Grid Data}} & \multicolumn{2}{c}{\textbf{Graph Data}}  \\
\cmidrule(lr){2-5} \cmidrule(lr){6-7}
& \multicolumn{2}{c}{\textbf{CrowdBJ}} & \multicolumn{2}{c}{\textbf{PopSH}} & \multicolumn{2}{c}{\textbf{TrafficNJ}} \\
\cmidrule(lr){2-3} \cmidrule(lr){4-5} \cmidrule(lr){6-7}
\textbf{Model} & \textbf{RMSE} & \textbf{MAE} & \textbf{RMSE} & \textbf{MAE} & \textbf{RMSE} & \textbf{MAE} \\
\cmidrule(lr){1-1} \cmidrule(lr){2-3} \cmidrule(lr){4-5} \cmidrule(lr){6-7}
HA & 0.343 & 0.232 & 0.165 & 0.1003 & 3.10 & 2.04 \\
ARIMA & 0.404 & 0.236 & 0.197 & 0.112 & 4.83 &  1.96  \\
\cmidrule(lr){1-1} \cmidrule(lr){2-3} \cmidrule(lr){4-5} \cmidrule(lr){6-7}
STResNet & 0.751 & 0.546 & 0.138 & 0.102 & - &  -\\
ACFM &0.200 & 0.141 & 0.078 & 0.0549 & - &  -\\
STNorm &0.198 & 0.132 & 0.065 & 0.042 & - &  -\\
STGSP & 0.229 & 0.157 & 0.057 & 0.040 & - &  -\\
MC-STL & 0.311 & 0.235 & 0.100 & 0.076 & - &  -\\
MAU & 0.256 & 0.166 & 0.125 & 0.081 & - & - \\
MIM & 0.298 & 0.214  & 0.126 & 0.079 & - &  -\\
SimVP & 0.213 & 0.148 & 0.069 & 0.048 & - & -\\
TAU & 0.196 & 0.135 & 0.063 & 0.044 & - &  -\\
PromptST & 0.171 & 0.099 & 0.069 & 0.0431 &- &  -\\
UniST &  0.172 & 0.106 &  0.055 &  0.036 & - & - \\
\cmidrule(lr){1-1} \cmidrule(lr){2-3} \cmidrule(lr){4-5} \cmidrule(lr){6-7}
STGCN &- &-&- & -&2.38 & 1.71 \\
DCRNN &- &- & -&- & 2.15 & 1.50\\
GWN & - &- & -&- & 2.23 & 1.58\\
MTGNN & - &- & -&- & 2.08 & 1.44\\
AGCRN &- &- & -&- & 2.54 & 1.86\\
GTS & - &- & -&- & 2.76 & 2.14\\
STEP & - &- & -&-& 2.19 & 1.55\\

\cmidrule(lr){1-1} \cmidrule(lr){2-3} \cmidrule(lr){4-5} \cmidrule(lr){6-7}
STID & 0.262 & 0.2037 & 0.055 & 0.038 & 2.45 & 1.64 \\
PatchTST & 0.291 & 0.189 & 0.108 & 0.062& 2.83 & 1.92\\
PatchTST(one-for-all) & 0.279 & 0.176 & 0.148&0.089 & 3.45 &  2.302\\
iTransformer & 0.249 & 0.154& 0.0722 & 0.0445 & 2.70& 1.79 \\
Time-LLM & 0.195 & 0.115 & 0.102 & 0.060 & 2.85 & 1.91\\
\cmidrule(lr){1-1} \cmidrule(lr){2-3} \cmidrule(lr){4-5} \cmidrule(lr){6-7}
\textbf{UniFlow} & 0.148 & 0.089 & 0.054 & 0.035 & 2.36 & 1.56 \\
\bottomrule
\end{tabular}}
\end{threeparttable}
\end{table}

\begin{table}[t!]
\caption{Performance comparison of long-term prediction on the remaining datasets in terms of MAE and RMSE. We use the average prediction errors over all prediction steps.}
\label{tbl:app_long}
\begin{threeparttable}
\resizebox{\columnwidth}{!}{
\begin{tabular}{ccccccc}
\toprule
& \multicolumn{2}{c}{\textbf{Grid Data}} & \multicolumn{4}{c}{\textbf{Graph Data}}  \\
\cmidrule(lr){2-3} \cmidrule(lr){4-7}
& \multicolumn{2}{c}{\textbf{TaxiNYC}} & \multicolumn{2}{c}{\textbf{TrafficSH}} & \multicolumn{2}{c}{\textbf{TrafficNJ}} \\
\cmidrule(lr){2-3} \cmidrule(lr){4-5} \cmidrule(lr){6-7}
\textbf{Model} & \textbf{RMSE} & \textbf{MAE} & \textbf{RMSE} & \textbf{MAE} & \textbf{RMSE} & \textbf{MAE} \\
\cmidrule(lr){1-1} \cmidrule(lr){2-3} \cmidrule(lr){4-5} \cmidrule(lr){6-7}
HA & 60.73 & 21.14 & 2.43 & 1.64 & 3.19 & 2.22\\
ARIMA & 86.61 & 27.68 & 2.05 & 3.89 & 4.11 & 2.83 \\
\cmidrule(lr){1-1} \cmidrule(lr){2-3} \cmidrule(lr){4-5} \cmidrule(lr){6-7}
STResNet & 36.51 & 18.42 & - & -& -& - \\
ACFM & 30.24 & 13.85&- & -& -& - \\
STNorm & 29.52 & 11.55 & - & -& -& - \\
STGSP & 25.59 & 11.06 & - & -& -& - \\
MC-STL & 37.85 & 24.45   & - & -& -& - \\
MAU & 30.55 & 10.56 & - & -& -& - \\
MIM & 56.47 & 57.01 & - & -& -& - \\
SimVP & 23.84 & 10.35 & - & -& -& - \\
TAU & 23.59 & 10.11 & - & -& -& - \\
PromptST & 29.11 & 10.02 & - & -& -& - \\
UniST &  22.69 &  7.85 & - & - & - & -\\
\cmidrule(lr){1-1} \cmidrule(lr){2-3} \cmidrule(lr){4-5} \cmidrule(lr){6-7}
STGCN & -& - & 3.41& 2.72&3.61&2.96 \\
DCRNN & -& - & 3.26 & 2.63 & 3.29 & 2.57\\
GWN & -& - & 3.61 & 2.96 & 3.51 & 2.76\\
MTGNN & -& - & 3.39 & 2.70 & 3.47 & 2.64 \\
AGCRN & -& - & 2.57 & 1.90 & 2.61 & 1.91\\
GTS & -& - & 3.12 & 2.46 & 3.61 & 2.83\\
STEP & -& - & 2.62 & 1.94 & 2.76 & 2.04\\

\cmidrule(lr){1-1} \cmidrule(lr){2-3} \cmidrule(lr){4-5} \cmidrule(lr){6-7}
STID & 22.04 & 9.59 & 2.24 & 1.56 & 2.52 & 1.72 \\
PatchTST & 31.79 & 10.43 & 2.19 & 1.38 & 2.56 & 1.69 \\
iTransformer & 32.83 & 11.09 & 2.18 & 1.47 & 2.56 & 1.75 \\
Time-LLM & 29.56 & 9.72 & 2.10 & 1.37 & 2.57 & 1.71\\
\cmidrule(lr){1-1} \cmidrule(lr){2-3} \cmidrule(lr){4-5} \cmidrule(lr){6-7}
\textbf{UniFlow} & 21.1 &  6.49 & 2.09 & 1.43 & 2.44 & 1.62 \\
\bottomrule
\end{tabular}}
\end{threeparttable}
\end{table}

We train the overall model in an end-to-end manner, aiming to minimize the mean squared error (MSE) loss between the predicted spatio-temporal data and the ground truth data. Specifically, we calculate the MSE only on the prediction horizon, even though the decoder outputs the complete spatio-temporal data. The loss function is formulated as follows:

\begin{align}
    \mathcal{L} = \frac{1}{N}\frac{1}{P} \sum_{i=1}^{N}\sum_{p=t_0+1}^{t_0+P} ||\hat{Y}_{i,p} - Y_{i,p}||^2 \nonumber
\end{align}

\noindent where $\hat{Y}_{i,p}$ is the predicted value, $Y_{i,p}$ is the ground truth,  $N$ is the number of samples in the dataset, and $P$ is the prediction horizon. We use the Adam optimizer to minimize the loss function and update the model parameters.

\textcolor{black}{We train the from scratch using all datasets simultaneously and directly evaluate its performance on the test sets of each dataset. This approach does not involve separate pre-training and fine-tuning stages. Instead, the model is trained end-to-end to perform the prediction task across all datasets. We use multiple datasets during training, and for each batch, we randomly sample from one of the datasets, allowing the model to jointly train across different data types.  
We use an A100 GPU for our experiments. The input data dimensions vary across different datasets, and these dimensions are provided in Appendix Table~\ref{tbl:append_data_grid} and Table~\ref{tbl:append_data_graph}. To adjust the batch sizes for different datasets, we employ the following strategy: $B=\frac{N}{K}$, where $K$ is the pre-determined number of iterations for each dataset and $N$ is the total number of training samples. We first set the batch size $B$ for the dataset with the largest spatial locations, ensuring optimal utilization of GPU memory. From this, we calculate $K$ and subsequently adjust the batch sizes for the remaining datasets accordingly. This approach allows for similar training iterations per epoch across different datasets while preventing GPU memory overflow.}

\subsubsection{\textbf{Evaluation Metrics}}
To comprehensively evaluate the performance of our model, we employ two widely-used metrics: Root Mean Squared Error (RMSE) and Mean Absolute Error (MAE).
The formulas for RMSE and MAE are as follows:

\begin{align}
\text{RMSE} = \frac{1}{N}\frac{1}{P} \sum_{i=1}^{N}\sum_{p=t_0+1}^{t_0+P} (\hat{Y}_{i,p} - Y_{i,p})^2\\
\text{MAE} = \frac{1}{N}\frac{1}{P} \sum_{i=1}^{N}\sum_{p=t_0+1}^{t_0+P} |\hat{Y}_{i,p} - Y_{i,p}|
\end{align}

\noindent where $\hat{Y}_{i,p}$ is the predicted value,  $Y_{i,p}$ is the actual value, $N$ is the number of samples in the dataset, and $P$ is the prediction horizon.

\subsection{Additional Results}\label{sup:result}

\subsubsection{\textbf{Prediction Performance}}

Table~\ref{tbl:app_short} reports short-term prediction results of the remaining datasets, and Table~\ref{tbl:app_long} reports long-term prediction results of the remaining datasets.

\subsubsection{\textbf{Efficiency Evaluation}}
\begin{table}[t!]
\caption{The information of training time and inference time.}
\label{tbl:traintime}
\centering
\resizebox{0.9\columnwidth}{!}{
\begin{tabular}{ccc}
\toprule
Model & train time all & inference time  \\
\hline
STGCN	& 17min	& 2s \\
DCRNN	& 77min	& 8s \\
GWN	& 16min & 1s \\
MTGNN	& 14min & 0.8s \\
AGCRN	& 21min	& 2s \\
GTS	& 126min & 17s \\
STEP & 177min 	& 27s \\
STResNet  & 5.7min & 0.6s \\
ACTM  & 56min & 0.9s\\
STNorm  & 46min & 5s \\
STGCP & 8min  &   4s \\
MC-STL  & 31min & 7s \\
MAU  & 82min & 13s  \\
MIM  & 84min & 14s  \\
TAU  & 22min & 6s \\
PromptST & 45min & 9s  \\
UniST  &  5h & 19s \\
STID & 10min &  4s  \\
PatchTST  & 33min & 5s  \\
iTransformer & 23min & 6s \\
Time-LLM & 382min &  5min \\
UniFlow & 6h (multiple datasets) & 10s \\
\bottomrule
\end{tabular}}
\end{table}
Table~\ref{tbl:traintime} reports model efficiency in terms of overall training time and inference time. Although the training time of UniFlow is significantly longer than the baseline models due to its inclusion of all datasets, it is important to note that training separate models for each dataset and summing the total training time results in comparable times between UniFlow and the baseline methods. Moreover, UniFlow achieves the best performance across all datasets with a single, unified model. This demonstrates the efficiency and effectiveness of UniFlow in delivering superior performance without the need for multiple specialized models.

\vfill

\end{document}